\documentclass[letterpaper, 10 pt, conference]{ieeeconf}
\IEEEoverridecommandlockouts
\usepackage{graphicx}
\usepackage{amsmath,amssymb}
\usepackage{booktabs}
\usepackage{multirow}
\usepackage{xcolor}
\usepackage{cite}
\usepackage{url}
\usepackage{cuted}
\usepackage{capt-of}

\newcommand{\method}{ReShoot}

\title{\LARGE \bf
ReShoot: Generative Visual Domain Randomization of Recorded\\ Robot Demonstrations for Visuomotor Policy Learning
}

\author{Chiyoung Kim$^{\dagger}$, Min Sung Choi$^{\dagger}$, Jinho Ju, Chanhoe Gu, Donghwan Hwang,\\
Wonseok Choi, Woongsun Jeon$^{*}$, and Minhyeok Lee$^{*}$%
\thanks{The authors are with Chung-Ang University, Seoul, Republic of Korea.
{\tt\small kimcy0829@cau.ac.kr, mscdavid0420@gmail.com, wlsgh20728@naver.com,
kum0100@cau.ac.kr, ghkd7545@cau.ac.kr, harry0695@cau.ac.kr, wjeon@cau.ac.kr,
mlee@cau.ac.kr}}%
\thanks{$^{\dagger}$These authors contributed equally.}%
\thanks{$^{*}$Corresponding authors.}%
}

\begin{document}

\maketitle
\thispagestyle{empty}
\pagestyle{empty}

\begin{strip}
\centering
\includegraphics[width=\textwidth]{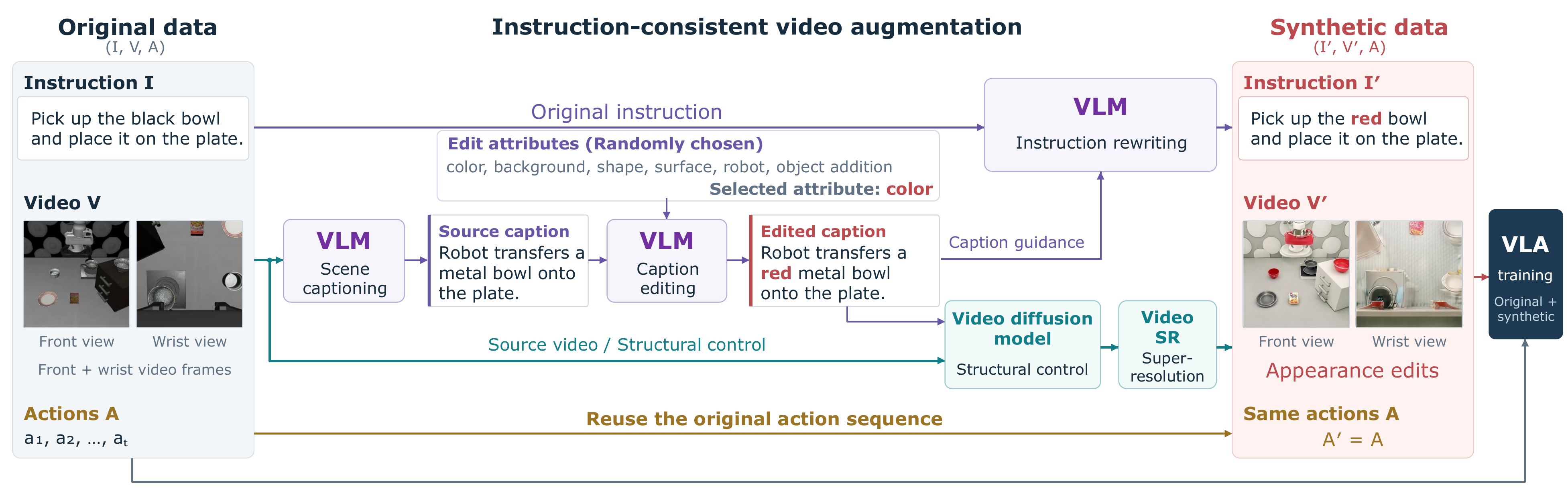}
\captionof{figure}{Overview of \method{}. A recorded demonstration provides the instruction, the third-person and wrist videos, and the actions. A vision-language model captions the scene and edits one attribute of the caption; a video generation module conditioned on the edge maps of the recorded frames redraws both views to match the edited caption, a super-resolution module restores sharpness, and the instruction is rewritten to match. The actions are reused unchanged.}
\label{fig:pipeline}
\end{strip}

\begin{abstract}
Imitation-learned robot policies are frequently overfit to the visual conditions present in their training demonstrations. Consequently, variations in object color or background appearance often induce substantial performance degradation. A common mitigation strategy is to acquire additional demonstrations in each novel visual context; however, this approach is resource-intensive, requiring repeated access to a robot, a controlled environment, and human operation for every appearance condition to be covered. We introduce \method, a framework that synthesizes visual diversity by re-rendering previously recorded demonstrations under altered appearances, thereby shifting the burden from data collection to generation. A vision–language model captions the scene, edits a targeted attribute (e.g., background, object color, or material), and an edge-conditioned video generator re-renders both camera views to match. The instruction is updated accordingly. The action sequence and proprioceptive trajectory are copied verbatim without relabeling, so each generated episode retains the recorded action and proprioceptive labels. On LIBERO, a policy trained on an equal mixture of recorded and re-rendered demonstrations matches the performance of recorded-only training (96.5\% vs.\ 96.9\%). Moreover, the mixed training set improves robustness to scene perturbations on LIBERO-Plus (85.5\% vs.\ 82.3\%). Across two physical robotic platforms, deploying \method{} with 43 and 100 pre-collected demonstrations increased the success rate on recolored objects from 0.0\% to 42.9\% and 47.5\%, respectively, while maintaining performance under the original recorded appearance.
\end{abstract}

\section{Introduction}

Consider a robotic system trained via imitation learning to grasp a black bowl and place it onto a plate. The system executes this manipulation task robustly under the training condition; however, when the black bowl is substituted with an otherwise identical red bowl, task performance degrades. Although the required kinematics and contact dynamics are unchanged, the learned policy fails due to limited generalization beyond the visual attributes represented in the training data. This sensitivity to distribution shift in perceptual inputs constitutes a well-documented failure mode in imitation learning. 

Visuomotor policies, from diffusion-based controllers~\cite{chi2023diffusionpolicy} to VLA models~\cite{brohan2023rt2,kim2024openvla,black2024pi0,kim2025oft}, are often highly sensitive to the visual conditions in their training demonstrations. A prevalent mitigation strategy is to acquire additional demonstrations in the target environment. However, each new data-collection episode entails access to a robotic platform, a suitably configured scene, and supervision by a human operator. For example, DROID required approximately 50 data collectors over a 12-month period~\cite{khazatsky2024droid}, and other large-scale datasets have been assembled through collection campaigns of comparable magnitude~\cite{brohan2023rt1,oxe2024,walke2023bridgedata}. The cost is driven by how many different scenes the demonstrations cover, not only by how many are recorded. Scaling studies show that the diversity of scenes and objects matters more for generalization than demonstration count~\cite{lin2025datascaling}. Since one collection setup covers only a narrow range of appearances, the effort grows with every appearance the policy must cover.

In simulation, this problem has a well-established solution. Domain randomization~\cite{tobin2017domain,peng2018sim2real,james2019rcan} keeps the trajectory and re-renders it under random textures and lighting, so that a single demonstration covers many appearances without additional robot time. Extending this idea to real recordings is not straightforward, because a recording exposes no renderer that could draw the same motion again with a red bowl. Research on real data has taken two other routes. One route edits single images with text-to-image diffusion models, inserting objects or replacing backgrounds~\cite{yu2023rosie,chen2023genaug,mandi2022cacti,bharadhwaj2024roboagent}; these methods work frame by frame with masks, so an edit does not propagate across a whole demonstration or between the two camera views. The other route generates new task videos with a video model and then infers the actions behind them~\cite{du2023unipi,ko2024avdc,jang2025dreamgen}; the visual diversity is large, but the action labels are inferred and not recorded.

Structure-conditioned video generation provides the renderer that real recordings lack. A video generation model that follows the structure of a recorded video~\cite{nvidia2025cosmostransfer,zhang2023controlnet} can redraw an entire demonstration under a new appearance while preserving the source motion structure, allowing the recorded actions to be reused without re-estimation. Changing the appearance while keeping the motion adds the coverage the policy needs. We take this third route, in which no action label is estimated.

Using a video model as a source of training data requires more than generating videos. Each generated frame must correspond to the action recorded at the same time step, while video models impose their own frame counts and resolutions; the third-person view and the wrist view must show the same new scene; when the recolored bowl appears, the instruction must name the new color, or the words no longer match the image. We present \method, a framework that meets these requirements and produces appearance variants of a recorded demonstration set (Fig.~\ref{fig:pipeline}). A vision-language model (VLM) describes the recorded scene and edits one attribute of that description, choosing among the background, an object's color or shape, the robot's appearance, the surface material, and an added static object. A video generation module conditioned on the edge maps of the recorded frames redraws the whole demonstration in both camera views at once, a super-resolution module restores sharpness, and the VLM rewrites the instruction. The actions and proprioceptive states of the recording are reused unchanged, so a new appearance requires a generation run and not a collection session.

We evaluate these episodes as training data on LIBERO, on LIBERO-Plus, and on two physical robots, under a fixed training budget that controls for the number of parameter updates. A mixture of recorded and re-rendered demonstrations matches recorded-only performance while improving robustness to perturbations. On the robots, policies trained on the existing demonstrations and their re-renderings succeed on recolored objects where recorded-only policies fail, with no further recording. In summary, this paper contributes
\begin{itemize}
\item a formulation of domain randomization for recorded robot data, in which each demonstration yields training episodes with new appearances while retaining its original action labels, reducing the need to collect separate demonstrations for each target appearance;
\item results showing that a policy drawing half of its training samples from re-rendered episodes matches recorded-only performance on LIBERO while improving robustness by 3.2 points on LIBERO-Plus;
\item measurements of the recorded structure that re-rendering preserves and of the appearance change it produces, together with a comparison of mixed and re-rendered-only training;
\item validation on two physical robots, where re-rendering existing demonstrations raises recolored-object success from complete failure to 42.9\% and 47.5\%.
\end{itemize}

\section{Related Work}

\textbf{Scaling demonstration data.} Large teleoperated datasets~\cite{oxe2024,khazatsky2024droid,walke2023bridgedata,brohan2023rt1} have made VLA pretraining possible~\cite{kim2024openvla,black2024pi0}, but each new scene still requires collection, and data scaling laws show that environment and object diversity matter more than demonstration count~\cite{lin2025datascaling}. MimicGen~\cite{mandlekar2023mimicgen} expands human demonstrations by adapting object-centric segments to new scene configurations, but the new trajectories must be executed to obtain observations, whereas \method{} changes the observations of trajectories that have already been executed.

\textbf{Generative augmentation for robot learning.} ROSIE~\cite{yu2023rosie} and GenAug~\cite{chen2023genaug} inpaint objects, distractors, and backgrounds into robot images with text-to-image diffusion; CACTI~\cite{mandi2022cacti} and RoboAgent~\cite{bharadhwaj2024roboagent} apply such augmentation frame by frame, GreenAug~\cite{teoh2024greenaug} replaces a green screen, and RoboEngine~\cite{yuan2025roboengine} generates backgrounds around a segmented robot. RoVi-Aug~\cite{chen2024roviaug} re-renders the robot or the viewpoint to transfer policies across embodiments. These methods operate on individual frames with masks or segmentation. \method{} generates whole videos from the structure of the recording, so one edit applies to the entire episode and to both camera views, and no mask is needed. Closer to our setting, Cosmos-Transfer1~\cite{nvidia2025cosmostransfer} conditions a video world model on edge, depth, and segmentation inputs, and RoboTransfer~\cite{liu2025robotransfer} enforces geometric and multi-view consistency in video diffusion to transfer manipulation policies. \method{} differs in its goal and in its labels. It addresses the cost of appearance-specific data collection, and it keeps the recorded action labels by aligning the generated frames with the recorded actions and rewriting the instruction to match the caption edit. We measure the value of this data against recorded data under a fixed training budget.

\textbf{Video models as policies and world models.} UniPi~\cite{du2023unipi} and AVDC~\cite{ko2024avdc} generate task videos and recover actions through inverse dynamics or dense correspondences, and DreamGen~\cite{jang2025dreamgen} labels generated videos with pseudo-actions. These approaches obtain their action labels by inference. \method{} constrains the generator to the recorded motion and keeps the recorded actions, so the training labels are the ones the robot executed.

\textbf{Domain randomization and robustness.} Randomizing rendering and dynamics in simulation transfers policies to the real world~\cite{tobin2017domain,peng2018sim2real,james2019rcan}, and LIBERO-Plus~\cite{fei2026liberoplus} shows that VLA models are brittle to controlled perturbations of LIBERO~\cite{liu2023libero} tasks. \method{} applies domain randomization to recorded trajectories with a generative model in place of the renderer.

\section{Method}
\label{sec:method}

\subsection{Problem Setting}
A recorded demonstration is $\tau=\big(\ell,\{(o^{f}_{t},o^{w}_{t},s_{t},a_{t})\}_{t=1}^{T}\big)$, with instruction $\ell$, third-person and wrist images $o^{f}_{t}$ and $o^{w}_{t}$, proprioceptive state $s_{t}$, and action $a_{t}$. \method{} produces
\begin{equation}
\tilde{\tau}=\big(\tilde{\ell},\{(\tilde{o}^{f}_{t},\tilde{o}^{w}_{t},s_{t},a_{t})\}_{t=1}^{T}\big),\quad
\tilde{o}^{v}_{1:T}=G\big(\tilde{c},\,E(o^{v}_{1:T})\big),
\label{eq:reshoot}
\end{equation}
where $v\in\{f,w\}$, $\tilde{c}$ is a scene description edited so that one visual attribute differs from the recorded scene, $E$ extracts per-frame edge maps, $G$ is the edge-conditioned video generation module, and $\tilde{\ell}$ is the instruction adapted to $\tilde{c}$. The trajectory is held fixed and only the appearance is randomized, which separates \method{} from methods that synthesize new motion and must recover the actions behind it. Because the actions are inherited rather than re-estimated, generation must preserve their correspondence with the frames, and edge conditioning is used to encourage this correspondence. Each recorded action $a_t$ therefore stays paired with the generated observation at the same time step. Appearance varies independently of trajectory and labels, and a single recording yields as many training episodes as appearances are generated. Generation therefore expands appearance coverage without additional robot interaction.

\subsection{Pipeline}
The pipeline consists of scene captioning, caption editing, edge-conditioned re-rendering, super-resolution, instruction rewriting, and episode assembly (Fig.~\ref{fig:pipeline}). The first two steps determine which attribute changes, the next two produce the new frames, and the last two apply the change to the instruction and to the stored episode.

\textbf{Scene captioning and caption editing.} A VLM first produces a short description of the third-person video, covering the scene, the objects, and the robot's motion. It then rewrites that description so that one attribute of the scene changes. We edit a caption of the real scene instead of writing a prompt from scratch, which applies the edit to the scene the caption describes. The prompt names six attribute types, namely background, object color, object shape, robot appearance, surface material, and an added static object, and asks the model to apply one type and to vary the choice across captions. Changing one attribute at a time keeps the edited caption close to the source scene and spreads the variation over the dataset. Leaving the choice to the model adapts each edit to the content of the scene. The edited description becomes the generation prompt $\tilde{c}$.

\textbf{Edge-conditioned re-rendering.} Edge maps are extracted from every frame of each view, and the video generation module, prompted with $\tilde{c}$, is conditioned on them through a structure-conditioning adapter. Edge maps are suited to this role for two reasons. Any recorded frame yields edge maps without a depth sensor or a segmentation model, and these maps constrain the contours of the robot and the objects while leaving color and texture free. Each view is generated separately, with the caption and the random seed reused across views to encourage a consistent appearance. The conditioning is kept strong through the late denoising steps, which preserves the structure while the appearance is synthesized. The caption edit names one attribute, but the module starts from noise and sees the recorded episode only through its edge maps, so the appearance changes beyond the edited attribute. Even a shape or added-object edit is rendered within the outline of the recorded scene, which leaves the geometry the actions depend on.

\textbf{Super-resolution.} At the policy input, the cropped frames are less sharp than recorded frames, because the generation module renders at a lower resolution. A video super-resolution module restores fine detail before the frames are resampled. Sharpness matters because a systematic blur would itself be a domain shift between the recorded and the re-rendered halves of the training data.

\textbf{Instruction rewriting.} The VLM rewrites the instruction to match the edited caption, so that ``pick up the black bowl \ldots'' becomes ``pick up the red metal bowl \ldots'' when the bowl is recolored (Fig.~\ref{fig:pipeline}). The rewrite is conditioned on the caption, which specifies the intended edit, and not on the generated frames, from which recovering the edit would add a second source of error. Rewriting aligns the instruction with the intended caption edit.

\textbf{Episode assembly.} The new episode is written in the same format as the recorded dataset. Only the images and the instruction change; every recorded label is kept. Because no label is recomputed, a re-rendered episode is used by the same training objective as a recorded one, with no change to the loss or the policy.

\subsection{Spatial and Temporal Alignment}
The generation module works at a different aspect ratio and frame count than the policy input, and any mismatch would misalign the generated observations and the recorded actions. Spatially, we keep the recorded aspect ratio by letterboxing each frame into the generator's canvas instead of stretching it, so the generator receives the edges of the recorded frame; the letterbox is cropped away afterwards. Temporally, the generation and super-resolution modules each accept only certain sequence lengths, so we pad each input by repeating its final frame and truncate the output back to $T$ frames. Generated frame $t$ and recorded action $a_t$ therefore carry the same index.

\subsection{Training with Re-rendered Demonstrations}
\label{sec:mixing}
Training draws from the two sources with weights that correct for their different sizes, so that each supplies half of every batch in expectation, and we use this equal proportion in all experiments. Generated frames only approximate the domain in which the policy will be evaluated, so recorded samples keep the policy anchored to that domain while the re-rendered ones broaden the appearances present in training. Section~\ref{sec:replace} evaluates training on re-rendered data alone to separate the two sources.

\section{Experimental Setup}
\label{sec:setup}

\begin{table}[t]
\caption{Models and settings used to instantiate the pipeline.}
\label{tab:settings}
\centering
\footnotesize
\setlength{\tabcolsep}{3pt}
\renewcommand{\arraystretch}{1.05}
\begin{tabular}{@{}p{0.24\columnwidth}p{0.73\columnwidth}@{}}
\toprule
Module & Model and settings \\
\midrule
Captioning, editing, rewriting & qwen3.6-flash~\cite{qwen36flash} \\
Video generation & Wan2.2-T2V-A14B~\cite{wan2025} at $832\times480$; UniPC sampler~\cite{zhao2023unipc}, 30 steps, guidance scale 3.0 \\
Structure conditioning & Holistically nested edge maps~\cite{xie2015hed} through a ControlNet~\cite{zhang2023controlnet,thedenk2025controlnet}; weight 1.1, active until 95\% of denoising; 1.0 and 80\% on the real robots, whose recorded frames carry more edge detail \\
Super-resolution & FlashVSR~\cite{zhuang2026flashvsr}, one step, upsampled to $1024\times1024$ and resampled to $256\times256$ \\
\bottomrule
\end{tabular}
\end{table}

\textbf{Benchmarks and data.} We use the four LIBERO suites~\cite{liu2023libero} (Spatial, Object, Goal, and Long) and LIBERO-Plus~\cite{fei2026liberoplus}, which perturbs LIBERO tasks along seven dimensions (layout, viewpoint, initial state, instruction, lighting, texture, and sensor noise). The recorded training data are the LIBERO training set used by OpenVLA~\cite{kim2024openvla}, with 1,693 episodes. For the re-rendered data we use 1,000 episodes, one variation of each, with suite and task proportions matching the recorded set. Table~\ref{tab:settings} lists the models that instantiate the pipeline.

\textbf{Training conditions.} We compare three policies trained under the same budget. Rec uses recorded data only, Mix uses an equal mixture of recorded and re-rendered data (Section~\ref{sec:mixing}), and Gen uses the re-rendered episodes only. OpenVLA-OFT~\cite{kim2025oft} is initialized from the base OpenVLA-7B checkpoint~\cite{kim2024openvla} and fully fine-tuned with the OpenVLA-OFT training configuration and its image augmentation. Training runs for 150,000 steps at batch size 16, with a learning rate of 0.00002 decayed tenfold after 100,000 steps.

\begin{figure}[t]
  \centering
  \includegraphics[width=\columnwidth]{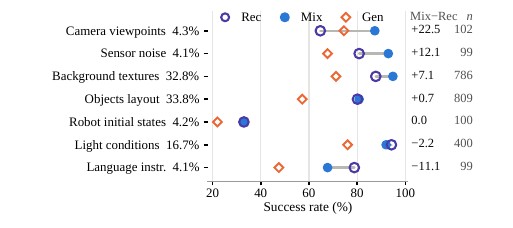}
  \caption{Success rate (\%) by LIBERO-Plus perturbation dimension. Percentages after the dimension names are their shares of the evaluated episodes; the right-hand columns give Mix minus Rec in percentage points and the number of episodes $n$.}
  \label{fig:simextra}
\end{figure}

\textbf{Evaluation.} The standard LIBERO benchmark uses 10 tasks with 50 trials each per suite from fixed initial states, 2,000 episodes in total. LIBERO-Plus holds about 2,500 perturbed tasks per suite. We group each suite into the six categories encoded in its task names, namely camera view, lighting, added object, table texture, task-board layout, and board level. From these we evaluate an evenly spaced sample of 2,395 perturbed episodes, one trial per task. The categories do not coincide with the seven perturbation dimensions of LIBERO-Plus, and Fig.~\ref{fig:simextra} gives the resulting share of each dimension. Re-weighting the per-dimension scores published for OpenVLA-OFT~\cite{fei2026liberoplus} by this sample gives 80.65\%, close to the 82.3\% we measure for Rec. Evaluation episodes are identical across conditions, so differences are tested with McNemar's exact test~\cite{mcnemar1947} on the discordant pairs.

\begin{table}[t]
\caption{Success rate (\%) on LIBERO and on the LIBERO-Plus perturbation sample. $\Delta$ is Mix minus Rec in percentage points, in bold when significant under the paired McNemar test ($p$).}
\label{tab:main}
\centering
\footnotesize
\setlength{\tabcolsep}{1.6pt}
\begin{tabular}{llccccc}
\toprule
 & Data & Spatial & Object & Goal & Long & All \\
\midrule
\multirow{5}{*}{LIBERO}
 & Rec & 97.4 & 99.4 & 98.6 & 92.2 & 96.9 \\
 & Mix & 96.8 & 99.4 & 97.2 & 92.4 & 96.5 \\
 & Gen & 71.8 & 89.2 & 91.6 & 55.2 & 77.0 \\
 & $\Delta$ & $-0.6$ & $0.0$ & $-1.4$ & $+0.2$ & $-0.5$ \\
 & $p$ & 0.70 & 1.00 & 0.19 & 1.00 & 0.45 \\
\midrule
\multirow{5}{*}{\shortstack[l]{LIBERO-\\Plus}}
 & Rec & 91.6 & 87.0 & 73.3 & 77.3 & 82.3 \\
 & Mix & 92.8 & 86.8 & 79.5 & 83.0 & 85.5 \\
 & Gen & 74.6 & 65.7 & 72.2 & 44.7 & 64.3 \\
 & $\Delta$ & $+1.2$ & $-0.2$ & $\mathbf{+6.2}$ & $\mathbf{+5.7}$ & $\mathbf{+3.2}$ \\
 & $p$ & 0.35 & 1.00 & $<0.001$ & 0.005 & $<0.001$ \\
\bottomrule
\end{tabular}
\end{table}

\begin{figure*}[t]
  \centering
  \includegraphics[width=\textwidth]{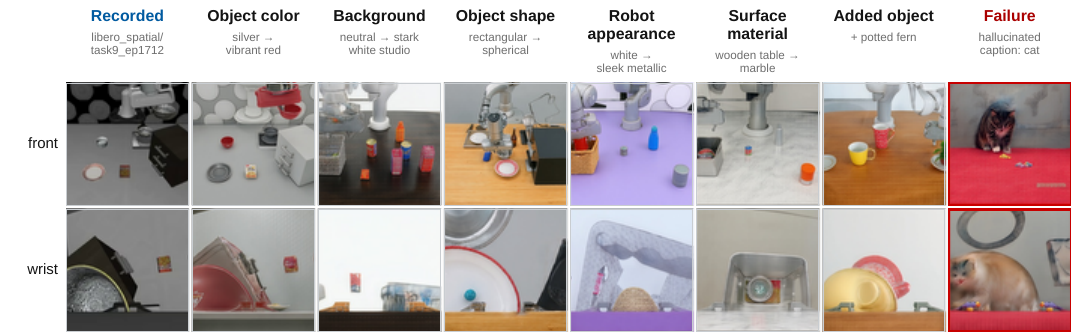}
  \caption{Re-rendered LIBERO demonstrations, third-person (top) and wrist (bottom) views. From left, a recorded frame, its re-rendering with the object color changed, one re-rendering for each of the other five attribute types as labeled, and a failure case.}
  \label{fig:qualitative}
\end{figure*}

\section{Results}
\label{sec:results}

\subsection{Standard LIBERO Performance under Mixed Training}
\label{sec:half}
Mix draws half of its training samples from re-rendered episodes and the other half from the recorded demonstrations. Mix reaches 96.5\% on standard LIBERO against 96.9\% for Rec, which trains on recorded data only, and the two policies match on every suite (Table~\ref{tab:main}). Under the same budget, Mix uses each recorded transition less than half as often as Rec.

Two properties of the re-rendered data, both following from the design of Section~\ref{sec:method}, explain this result. The action labels are those of the recording, so every re-rendered sample supervises the recorded motion, and the structure that the motion follows survives re-rendering. The per-frame edge maps of recorded and re-rendered episodes have a Pearson correlation of 0.80 for the third-person view and 0.70 for the wrist view. Both values are more than twice the 0.30 measured between wrist frames from unrelated time steps, and below the 0.96 between adjacent recorded frames, so much of the edge structure is preserved. The wrist view is the more demanding measurement, since it moves with the gripper and observes the region in which the manipulation occurs. At each step the policy sees the outline of the recorded scene under a new appearance, paired with the action that was recorded there. Super-resolution raises the sharpness of the generated frames. Measured as the Laplacian variance of the luma, it raises the sharpness of generated frames to the level of recorded frames, by $2.67\times$ for the third-person view and $2.19\times$ for the wrist view, while leaving pixel values almost unchanged. A re-rendered sample therefore retains the recorded action labels, and under this training budget, Mix matches recorded-only performance on this benchmark while drawing half of its training samples from re-rendered episodes. Importantly, this result also provides task-level evidence that the re-rendered observations remain sufficiently aligned with the inherited actions: replacing half of the training samples with re-rendered episodes does not measurably degrade performance on standard LIBERO.

\subsection{Robustness Gains Extend Beyond the Edited Attributes}
\label{sec:robust}
On LIBERO-Plus, Mix outperforms Rec by 3.2 points (85.5\% vs.\ 82.3\%). The gain is concentrated on the Goal and Long suites; on Spatial and Object, the two policies perform at the same level.

The improvements are not limited to the dimensions that caption editing targets (Fig.~\ref{fig:simextra}). The largest gain, 22.5 points, occurs for changes in camera viewpoint, and sensor noise improves by 12.1 points, even though the pipeline does not modify either of these aspects. Background texture, which the pipeline does modify, rises by 7 points; since this dimension makes up roughly a third of the samples, it accounts for most of the overall gain. Lighting shows no improvement despite frequent caption edits to lighting-related words, and the language-instruction dimension (affected by the instruction-rewriting step) drops the most. This behavior is characteristic of domain randomization rather than attribute-specific augmentation and follows from how the frames are generated. Conditioning fixes the scene layout while leaving other visual factors unconstrained (Section~\ref{sec:method}), and the caption edit merely specifies a direction of change that the module then applies across the entire scene. Pixel-level comparisons further support this interpretation. On the 0–255 scale, recorded versus re-rendered frames differ by 81 in the third-person view and 47 in the wrist view, over twenty times larger than the difference between consecutive recorded frames. This discrepancy is at least as large in parts of the image that contain none of the edited object. As a result, the policy experiences the same trajectory under many different appearances. The robustness it develops therefore transfers to perturbations that preserve scene structure, such as camera shifts or added sensor noise. 

An explanation for the lighting exception is that a lighting term in the caption does not alter rendered illumination as directly as a texture term alters rendered texture. The drop on language instructions has a separate cause: rewriting shifts training language away from the phrasing used at evaluation time, reducing accuracy on that dimension, even though the same rewriting helps the policy follow a modified color word on Robot~A. Since the robustness in the other dimensions is not tuned to any specific anticipated deployment perturbation, the attributes to edit need not be specified ahead of time. ReShoot does not require deployment perturbations to be specified in advance, unlike augmentation strategies crafted to address a known distribution shift.

\subsection{Effect of Re-rendered Demonstrations}
\label{sec:replace}
Gen reaches 77.0\% on LIBERO and 64.3\% on LIBERO-Plus, 18 to 20 points below Rec on both benchmarks, and both gaps are significant. Two properties of the generated data account for this gap. The instruction was rewritten in 26 to 48\% of the episodes, depending on the suite, while the benchmark evaluates with the original wording, so a policy trained on re-rendered data alone has learned a different mapping from words to tasks. The pixels, in turn, only approximate the rendering on which the benchmark evaluates. They also enter training without quality filtering, since the 20.5\% of episodes whose caption edit left the description unchanged and the 4.5\% whose caption named no robot all remain in the training set that produces the gains above. The same episodes form half of the data of Mix without costing accuracy there, because the two sources cover different parts of the problem. The recorded half supplies observations from the evaluation domain, and the re-rendered half supplies the appearance variation that the recorded set lacks. Gen exceeds Rec on camera viewpoints (Fig.~\ref{fig:simextra}), the dimension on which the mixture gains most, so on that dimension a policy trained without any recorded frame already carries the robustness that re-rendering adds. Object color (37\%), surface material (28\%), and background (23\%) account for most of the edits, while edits to shape, added objects, and robot appearance are rare, so re-rendering varies the appearance of the scene and leaves its layout as recorded.

\subsection{Real-Robot Validation}
\label{sec:real}
The simulation results motivate two empirically testable outcomes on physical robotic platforms: (i) preservation of performance under the appearance conditions observed during data collection, and (ii) robustness under an appearance shift. We evaluate both criteria on two distinct manipulators (Table~\ref{tab:real}). For each platform, the recorded-only policy is trained exclusively on the recorded demonstrations, whereas the augmented policy (denoted +\method{} in Table~\ref{tab:real}) is trained on the same demonstrations augmented with their corresponding re-rendered episodes. The augmented policies use the same recorded trajectories as the recorded-only policies, with additional re-rendered observations that expand appearance coverage. Both policies instantiate OpenVLA-OFT. We operationalize the appearance shift via object color, since color can be manipulated in a controlled manner and its inclusion in the re-rendering distribution can be verified directly from the synthesized frames. The first experiment introduces a distractor object and evaluates color-conditioned object selection; the second comprises four tasks and assesses invariance to color variations that are task-irrelevant.

\begin{figure}[t]
  \centering
  \includegraphics[width=\columnwidth]{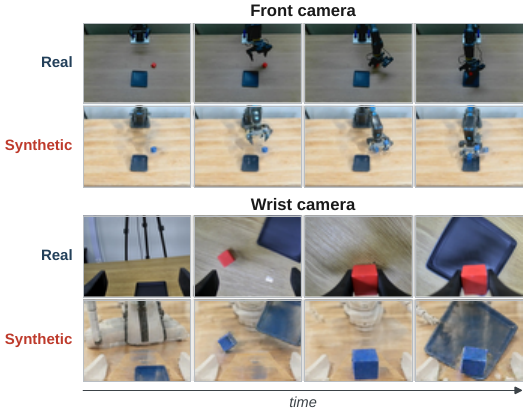}
  \caption{Robot~A. A recorded episode (rows 1 and 3) and its re-rendering (rows 2 and 4); third-person view above, wrist view below.}
  \label{fig:real_cube}
\end{figure}

\textbf{Robot A, cube pick-and-place with a distractor.} An AgileX PiPER manipulator equipped with a fixed overhead camera and a wrist-mounted camera is tasked with grasping a red cube and placing it onto a tray. The recorded dataset comprises 43 demonstrations spanning the arm’s reachable workspace on the tabletop. The re-rendered dataset augments each demonstration with one additional episode in which the target cube is blue and the natural-language instruction explicitly specifies that color (Fig.~\ref{fig:real_cube}). Both learned policies are evaluated from matched initial conditions. Under the original instruction, the policy trained solely on recorded data achieves a 70.0\% success rate over 20 trials, whereas the augmented policy achieves 50.0\%; with 20 trials, this apparent difference is not statistically well resolved. 

\begin{figure}[t]
  \centering
  \includegraphics[width=\columnwidth]{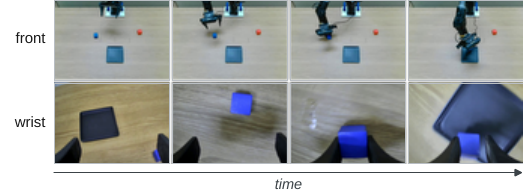}
  \caption{Robot~A. Rollout of the augmented policy under the appearance shift; third-person view above, wrist view below.}
  \label{fig:realextra}
\end{figure}

Under the distribution shift, both a red and a blue cube are present simultaneously on the table, a configuration not observed during training, while the input instruction refers to the blue cube. Evaluating success with respect to the instructed object, the recorded-only policy achieves 0.0\% success (0/21), whereas the augmented policy achieves 42.9\% success (9/21). In all successful trials, the augmented policy delivers the instructed blue cube (Fig.~\ref{fig:realextra}), indicating a substantial performance difference.

\begin{table*}[t]
\caption{Real-robot success (successes out of trials). Rec.\ denotes training on recorded demonstrations only, and +\method{} adds their re-renderings. $p$ is from McNemar's test on matched initial positions for Robot~A and from Fisher's exact test for Robot~B, whose placements are matched across conditions but not paired trial by trial.}
\label{tab:real}
\centering
\footnotesize
\setlength{\tabcolsep}{10pt}
\begin{tabular}{lcccccc}
\toprule
\multicolumn{7}{l}{\textit{Robot A, cube pick-and-place, recorded with a red cube}} \\
 & \multicolumn{3}{c}{Red cube} & \multicolumn{3}{c}{Blue cube with red distractor} \\
\cmidrule(lr){2-4}\cmidrule(lr){5-7}
Success definition & Rec. & +\method{} & $p$ & Rec. & +\method{} & $p$ \\
\midrule

Instructed cube & 14/20 & 10/20 & 0.34 & 0/21 & \textbf{9/21} & 0.004 \\
\midrule
\multicolumn{7}{l}{\textit{Robot B, four tasks, recorded with black objects}} \\
 & \multicolumn{3}{c}{Black objects} & \multicolumn{3}{c}{Recolored objects} \\
\cmidrule(lr){2-4}\cmidrule(lr){5-7}
Task & Rec. & +\method{} & $p$ & Rec. & +\method{} & $p$ \\
\midrule
Cylinder on cube         & 5/10 & 6/10 & & 0/10 & 6/10 & \\
Slipper on shelf         & 7/10 & 7/10 & & 0/10 & 4/10 & \\
Hollow cylinder on hook  & 5/10 & 6/10 & & 0/10 & 5/10 & \\
Cup on teacup            & 4/10 & 4/10 & & 0/10 & 4/10 & \\
Total                    & 21/40 & 23/40 & 0.82 & 0/40 & \textbf{19/40} & $<0.001$ \\
\bottomrule
\end{tabular}
\end{table*}

\begin{figure}[t]
  \centering
  \includegraphics[width=\columnwidth]{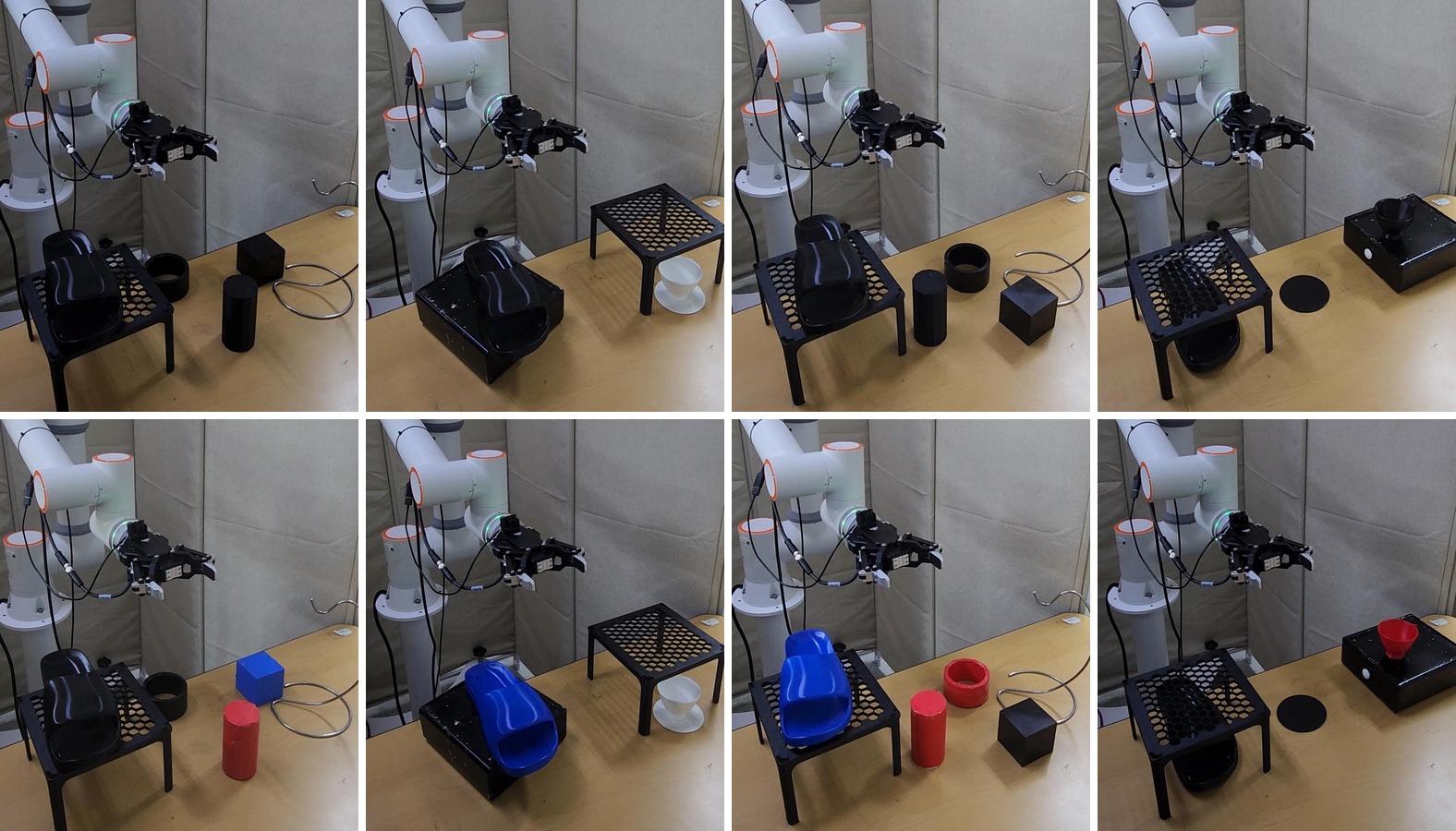}
  \caption{Robot~B. Initial scenes of the four tasks with black objects (top) and recolored objects (bottom).}
  \label{fig:real_tasks}
\end{figure}

\textbf{Robot B, four tasks under an object-color shift.} A FAIRINO FR5 arm with a third-person and a wrist camera performs four placement tasks, namely (i) placing a cylinder on a cube, (ii) a slipper on a shelf, (iii) a hollow cylinder on a hook, and (iv) a cup on a teacup. All 100 recorded demonstrations only use black objects (Fig.~\ref{fig:real_tasks}). To test whether a single recorded color can serve as the source for several, \method{} re-renders each demonstration seven times with the objects recolored in red, blue, green, white, and other colors. The resulting 700 episodes are trained on together with the 100 recorded ones. With black objects the two policies perform at the same level (52.5\% vs.\ 57.5\%). Under the evaluation with novel colors, the policy trained solely on recorded data fails to solve any of the 40 evaluation trials spanning four tasks (0\% success). By comparison, the augmented policy attains a 47.5\% success rate. Notably, the action sequences in these experiments originate from episodes recorded with black objects; thus, the robot was not provided with real demonstrations containing the novel colors. Instead, the policy’s representations of red, blue, green, and white objects were acquired from synthetic demonstrations by \method{}, and this learned information transferred to real-robot execution. The substantial real-robot success obtained using inherited action labels further indicates that ReShoot preserves the observation--action correspondence required for policy learning, as systematic misalignment would be expected to substantially impair physical execution.

\section{Discussion}
\label{sec:discussion}
In both simulation and real-robot settings, re-rendered demonstrations complement recorded demonstrations, allowing generative augmentation to reduce the need for additional appearance-specific data collection. Re-rendering preserves the original supervision (labels) while augmenting visual context, yielding domain-randomization-like robustness gains. We see better performance under camera-view and sensor-noise variations not explicitly altered by the pipeline, and on real robots, higher success under color shifts absent from the recorded dataset.

We hypothesize that showing the same trajectory in multiple visual instantiations reduces dependence on superficial appearance cues while preserving invariant structure. Under this view, diversity across edits matters more than any single attribute, so we let the VLM vary attributes across episodes. Instruction rewriting follows the same idea for language: it moves training utterances from a single benchmark phrasing toward references to objects actually present. On LIBERO-Plus, this shift reduces accuracy when evaluation retains the original wording, whereas it produces the desired instructed-object behavior on Robot~A.

The real-robot experiments show that even a single visual attribute change can be sufficient to break a policy trained from recorded demonstrations. In both platforms, changing only the object colors from those observed in the real demonstrations was sufficient for the recorded-only policies to fail every evaluation trial under the shifted condition, despite the underlying manipulation skills and task dynamics remaining unchanged. This result demonstrates that appearance variation is not merely a cosmetic nuisance: even a simple color change constitutes a severe distribution shift for visuomotor policies trained on visually narrow demonstrations. ReShoot directly targets this failure mode by re-instantiating already-recorded trajectories under alternative appearances while retaining their original action supervision. Accordingly, the augmented policies recovered substantial success under the color shifts (42.9\% and 47.5\% on the two platforms) without requiring additional real demonstrations in those appearance conditions. These results show that generative re-rendering improves the robustness of existing demonstration datasets against visually simple but behaviorally consequential distribution shifts.

\section{Limitations}
\label{sec:limitations}
\textbf{Structure preservation.} The edge conditioning constrains geometry but not appearance (Section~\ref{sec:method}) and does not enforce pixel-level correspondence, so object and gripper contours can drift from the recorded geometry. Depth or segmentation conditioning can tighten this constraint further.

\textbf{Text-only relabeling.} Instruction rewriting is conditioned on the edited caption and not on the generated frames, so a disagreement between caption and video reaches the rewritten instruction. Verifying the generated video before relabeling is a direct next step.

\textbf{Relation to simple photometric augmentation.}
Conventional photometric augmentations such as color jitter are not designed to provide the type of variation targeted by ReShoot. Such transformations perturb low-level image statistics globally and independently of task semantics, whereas ReShoot modifies semantically meaningful scene attributes while maintaining consistency with the recorded trajectory and corresponding language instruction. For example, ReShoot can alter a task-relevant object's appearance, surface material, background, or surrounding scene content while preserving the recorded action supervision; these transformations cannot in general be expressed by global photometric operators. A comparison centered on color jitter would therefore evaluate a considerably narrower augmentation regime rather than the central capability studied here. Our experiments instead examine whether generatively re-rendered demonstrations can replace additional appearance-specific data collection while preserving task performance and improving robustness.

\section{Conclusion}
\method{} gives recorded robot demonstrations a new appearance with an edge-conditioned video generation module while keeping their actions and states unchanged, which expands the visual diversity of an existing demonstration set without collecting demonstrations for each new appearance. On LIBERO, a policy trained on an equal mixture of recorded and re-rendered demonstrations performs at the level of recorded-only training and is more robust under perturbation. On two physical robots, re-rendering existing demonstrations yields success on object colors for which recorded-only policies score zero. \method{} adds visual diversity after demonstrations have already been recorded, reducing the need to anticipate appearance variation during data collection. Automatic checking of generated episodes and richer geometric conditioning can extend \method{} further, toward making video generation a practical component of robot data pipelines.

\end{document}